\def\year{2021}\relax
\documentclass[letterpaper]{article} % DO NOT CHANGE THIS
\usepackage{aaai21}  % DO NOT CHANGE THIS
\usepackage{times}  % DO NOT CHANGE THIS
\usepackage{helvet} % DO NOT CHANGE THIS
\usepackage{courier}  % DO NOT CHANGE THIS
\usepackage[hyphens]{url}  % DO NOT CHANGE THIS
\usepackage{graphicx} % DO NOT CHANGE THIS
\usepackage{natbib}  % DO NOT CHANGE THIS AND DO NOT ADD ANY OPTIONS TO IT
\usepackage{caption} % DO NOT CHANGE THIS AND DO NOT ADD ANY OPTIONS TO IT
\usepackage{amsmath}
\usepackage{amsfonts}
\usepackage{cleveref}
\usepackage{multirow}
\usepackage{booktabs}
\usepackage[table,xcdraw]{xcolor}
\usepackage[switch]{lineno}

\title{CNN Profiler on Polar Coordinate Images for Tropical Cyclone Structure Analysis}
\iftrue
\author {
        Boyo Chen,
        Buo-Fu Chen,
        Chun-Min Hsiao\\
}
\affiliations {
    National Taiwan University \\
    boyochen722@gmail.com, bfchen777@gmail.com, a0979155270@gmail.com
}
\fi
\begin{document}
%\linenumbers

\maketitle

\begin{abstract}
Convolutional neural networks (CNN) have achieved great success in analyzing tropical cyclones (TC) with satellite images in several tasks, such as TC intensity estimation. In contrast, TC structure, which is conventionally described by a few parameters estimated subjectively by meteorology specialists, is still hard to be profiled objectively and routinely. This study applies CNN on satellite images to create the entire TC structure profiles, covering all the structural parameters. By utilizing the meteorological domain knowledge to construct TC wind profiles based on historical structure parameters, we provide valuable labels for training in our newly released benchmark dataset. With such a dataset, we hope to attract more attention to this crucial issue among data scientists. Meanwhile, a baseline is established with a specialized convolutional model operating on polar-coordinates. We discovered that it is more feasible and physically reasonable to extract structural information on polar-coordinates, instead of Cartesian coordinates, according to a TC’s rotational and spiral natures. Experimental results on the released benchmark dataset verified the robustness of the proposed model and demonstrated the potential for applying deep learning techniques for this barely developed yet important topic.
\emph{Github link and dataset for the paper will be provided after the double blind review.}
\end{abstract}

\section{Introduction}
\label{sec:intro}

A tropical cyclone (TC), also called hurricane or typhoon, is a kind of rotating storm formed on tropical oceans; it is characterized by a low-pressure center (i.e., the “eye”), eyewall associated with deep convective clouds and strong winds, and spiral rainbands outside of the eyewall. 
This severe weather system often causes serious damage to human society due to gusty winds, torrential rainfall, high waves, and storm surges.

Although the improvement of TC forecasting in recent years ensures fairly well prediction of the track and primary rainfall distribution of a TC, there is still room for improvement in the ability to predict TC structure \citeauthor{knaff2016using,sampson2015consensus,sampson2018tropical}. 
Moreover, the TC structure, in terms of its 2-D surface wind fields, is closely related to the potential TC damage, the area affected by gale-force winds, and the magnitude of storm surges \cite{powell2007tropical}. 
Moreover, a better TC structure analysis served as the initial data of numerical weather prediction models is critical to improving the prediction accuracy regarding TCs \cite{tallapragadacoauthors,bender2016impact}.

It is not easy to accurately analyze the structure of a TC, noting that TCs spend most of their lifetime on the open ocean, where meteorological observation is severely limited. 
Therefore, meteorologists strongly relay on satellite remote sensing to estimate TC surface wind field, TC radial wind profile, and structural parameters (e.g., intensity, the radius of maximum wind, size; please refer to \cref{sec:related}). 

\begin{figure}[ht]
\centering
\includegraphics[width=0.9\linewidth]{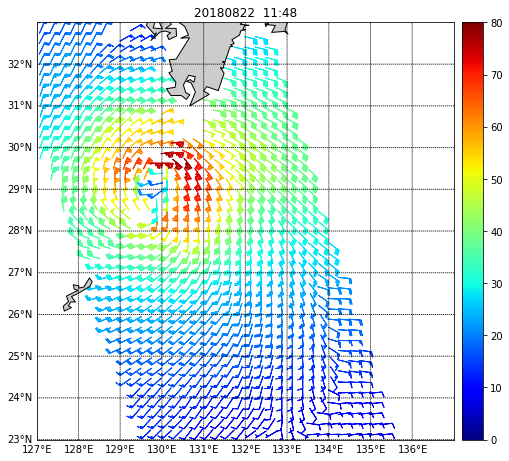}
\caption{The ASCAT surface winds (colored vectors, kt) observation of Typhoon Soulik (2018) at August 22 2018 11:48 UTC. The raw ASCAT data can be download from http://www.remss.com/.}
\label{fig:ASCAT}
\end{figure}

The most straightforward way to analyze TC structure is using the spaceborne radar surface wind observation (\cref{fig:ASCAT}), such as Advance Scatterometer (ASCAT, \cite{figa2002advanced,knaff2011automated}). 
Although ASCAT provides high-quality surface wind observation outside of the TC inner-core (i.e., larger than approximately 80-150 km radius), ASCAT only provides two scans of a TC per day. Sometimes, only a portion of the TC is observed due to ASCAT’s limited swath width.

To estimate TC structure at a higher frequency, other kinds of satellite observations have to be used, such as microwave sounders on low-Earth-orbit satellites \cite{knaff2011automated,demuth2004evaluation,wimmers2019using} and images from geostationary satellites \cite{velden2006dvorak,knaff2014objective,chen2019estimating}.
Infrared images that observe cloud features associated with a TC can be used to estimate several important TC structural parameters, including intensity  and size. 
For instance, Knaff et al. (2014) developed a TC size estimation technique based on feature engineering. Their model utilized principal component analysis of the azimuthally-averaged radial profile of the infrared brightness temperature and linear regression to estimate TC size.
With the structural parameters retrieved from satellite images, the TC radial wind profile can be constructed by a physically-based parametric wind model \cite{knaff2016using,morris2017determining}. 
However, there is some difficulty in using such a sample parametric model to analyze the TC structure.   

Although satellite remote sensing provides various observational data for TC structure analysis, the conventional statistical methods face limitations in analyzing multi-channel, high-dimensional, and temporal-spatially heterogeneous satellite data. 
Meanwhile, deep learning has achieved great success in analyzing satellite remote-sensing images of tropical cyclones (TC), such as TC intensity estimation \cite{chen2019estimating}, predicting TC intensification \cite{bai2019attention,yang2020long}, and anticipating TC formation \cite{matsuoka2018deep}. 
In these studies, Convolutional Neural Network (CNN,\cite{krizhevsky2012imagenet}) successfully extracted features that are difficult to be quantified before, form high-dimensional data, and use them to complete the classification or regression work.
These deep learning models provided more efficient, stable, and objective guidance for TC forecasting, and their performance is comparable but not significantly exceed the state-of-the-art meteorological techniques.

The goal of this study is to explore the potential of deep learning in this necessary but not well-tackled topic in meteorology.
To remove the dependencies of any sample parametric model and analyze TC structure directly with satellite images, we construct and release a new benchmark dataset, in which TC wind profiles were constructed based on meteorological domain knowledge to provide valuable data labels.

Furthermore, we propose a novel specialized CNN model operating on polar coordinates. Several different loss function compositions and model structures are explored and discussed in the following section. By properly designing our model, the experimental result show the promising future for the deep leaning techniques in this new topic.

This paper is organized as follows. Section 2 describes the definitions of TC structure and structural parameters, and how meteorologist conventionally estimates it. 
Section 3 describes our new-released dataset: Dataset of Tropical Cyclone Structural Analysis. 
Section 4 proposes the CNN architecture on polar coordinates, suitable for processing TC satellite imagery that is rotationally invariant. 
Section 5 includes the experiment results. 
Section 6 is the conclusion.

 \section{Background Knowledge}
\label{sec:related}

\subsection{Definition of TC structure and structural parameters}

As a cyclonically rotated system, a TC’s structure is usually referred to as the characteristics of the storm-centered surface wind field, which is closely related to the potential TC impacts.
But noting that TC is fairly axis-symmetric with respect to the center and has the tangential wind components much larger than the radial wind components (\cref{fig:ASCAT}), it is more practical to describe TC’s structure by the azimuthally-averaged radial surface wind profile \cite{holland1984dynamics,knaff2016using}.

With such a profile (fig. 2a, green line, as an example), several important structural parameters can be defined. 
TC intensity ($V_{max}$) is conventionally defined as the maximum wind near the TC center, and the radius of the maximum wind (RMW) indicates where $V_{max}$ occurs. 
TC’s size is usually defined as the radial extent from the center of certain wind thresholds, such as 5, 34, 50, or 64 kt. 
The 34 kt winds radius ($R_{34}$) is considered the most practical TC size parameter as it strongly relates to a TC’s impact. 
These three parameters are the most critical parameters to be estimated in real-time TC forecasting in operational weather prediction centers.
Furthermore, previous studies \cite{chan2012size,weatherford1988typhoon} have shown that the intensity is not strongly related to the size. 
This implies that knowing the intensity, which is the easiest one to estimate among the three parameters, is not sufficient to determine the structure of a TC. 

\subsection{Conventional method to estimate TC structure}

A space-borne scatterometer (e.g., ASCAT) provides high-quality surface wind observation \cite{figa2002advanced,knaff2011automated}. 
ASCAT is a C-band radar that measures ocean roughness and uses it to retrieve surface winds under approximately 30 m/s. 
Thus, the subjective analysis by forecasters based on scatterometer observation is considered one of the best metrics for analyzing TC structure \cite{sampson2017tropical,sampson2018tropical}.
However, the sampling frequency (twice pre-day) of ASCAT is not enough for operational TC structure analysis, which better has a higher frequency (less than six hour). 

A method, "Multiplatform Tropical Cyclone Surface Winds Analysis" (MTCSWA) \cite{knaff2011automated}, to estimate TC surface wind fields every 6 h utilized observation from multiple satellite platforms and satellite-based wind retrieval techniques. 
MTCSWA uses a variational data-fitting method to merge satellite observations that are temporal-spatially heterogeneous.
Although this method produces wind estimates with generally smaller errors than single raw input data, the analysis quality may be unstable when some important input data (i.e., ASCAT) are not available. 

The other approach to estimate an axis-symmetric TC structure is to estimate key structural parameters. 
Several studies \cite{knaff2014objective} have applied CNN for estimating TC intensity utilizing satellite imagery. 
On the other hand, Knaff et al. (2014) related the storm-centered satellite infrared imagery to TC size, in terms of the radius of azimuthally-averaged 5-kt winds. 
They created a multivariable linear regression equation based only on the first three principal components of the azimuthally-averaged radial profile of the infrared brightened temperature.

Estimating these structural parameters strongly relies on extracting high-level features from satellite images. 
However, some of the methods are subjective and depended on weather forecasters’ human intelligence; other objective statistical methods can only handle limited features or predicters. 
As there may be a great potential to extract more useful features by deep learning, we are motivated to apply deep learning for estimating not only a single structural parameter but, hopefully, the entire radial surface wind profile (i.e., the profiles as shown in fig. 2a).

\section{Dataset: TCSA}
\label{sec:data}
A new dataset for Tropical Cyclone Structure Analysis (TCSA) is released along with this research. TCSA can be used to develop deep learning models that estimate TC structural parameters (e.g., intensity, size, size asymmetry) and, more importantly, the axis-symmetric wind profile of the storm. \textit{Link to the dataset repository will be provided here after the double blind review.}

As an extension of another open dataset – Dataset of Tropical Cyclone for Image-to-intensity Regression (TCIR) \cite{chen2018rotation}, TCSA contains \textbf{76835 TC images}, collected from 2004 to 2018, covering 1407 TCs in every basin over the globe. For each TC, images are collected once per 3 hours. \textbf{The center of the TCs are always placed at the center of the images}.

\subsubsection{4 satellite channels} are included in every images: (1) infrared, (2) water vapor, (3) visible light channel, and (4) passive micro-wave rain rate (\Cref{fig:coo_example}(a)). 

\subsubsection{4 labels} are provided, including (1) intensity ($V_{max}$, defined as the maximum wind velocity), (2) size ($R_{34}$, defined as the mean of radii of 34-knot wind in the four quadrants, in kilometer), (3) radius of maximum wind speed (RMW), and, most important of all, (4) the wind profile.

\begin{figure}[t]
\centering
\includegraphics[width=\linewidth]{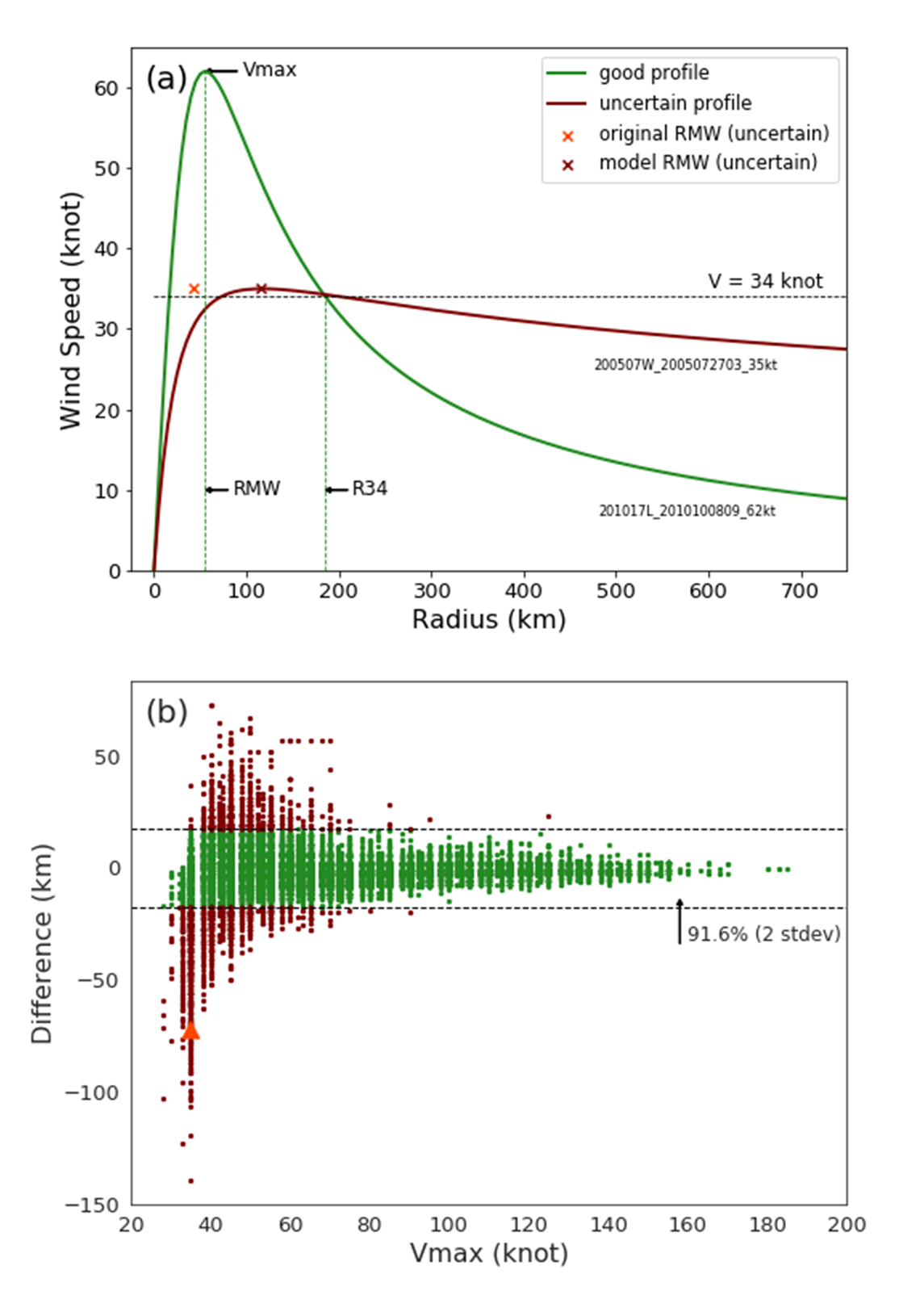}
\caption{(a) A good wind profile and an uncertain wind profile. For the good profile (green), $V_{max}$, RMW, and $R_{34}$ are indicated. For the uncertain profile (brown), the best-track RMW (orange "x") and the calculated RMW (brown "x") differ to each other. (b) The scatter plot of RMW difference vs. $V_{max}$. The horizontal lines indicate the interval of 2 standard deviations in Y-axis. The red triangle indicates the position of the uncertain sample as shown in (a). }
\label{fig:dataset}
\end{figure}

\subsection{Wind Profile Label}
\label{sec:profile_calculation}
In the TCSA, we apply the parametric wind model \cite{morris2017determining} to calculate the TC wind profile for every available data. The radial wind profile of a TC can be described by

$$V(r)=  \frac{2r(R_mV_{max} +\frac{1}{2} f R_m^2))}{(R_m^2+ar^b )}-\frac{fr}{2}$$

where $R_m$ is the radius of maximum wind speed (RMW), $V_{max}$ is the maximum wind speed, $r$ is the radial distance from the storm center, and $f$ is the Coriolis parameter. 
Here, we use $R_m$ , $V_{max}$, and $R_{34}$ to approach the most possible wind profile of the TC, with parameters a and b calculated by iteration. This wind model assumes that the TC is symmetry, and the adjustment in a and b allows fitting the wind speed profile better. According to meteorological domain knowledge, $R_{34}$ and $V_{max}$ are fixed in our calculation because of their higher reliability than that of RMW. $R_{34}$ and $V_{max}$ are also more critical in accessing TC impact in operational TC forecast. Consequently, RMW is allowed to be adjusted during the iteration. However, sometimes there might be a large difference between the original RMW and calculated RMW (fig. 2a), especially for weaker TCs. In such cases, we would question the correctness of the calculated profile.

Although we collected over 76000 images, \textbf{only 46\%} data, 35310 images, can be equipped with a valid wind profile label. This is usually because a sample's $R_{34}$ does not exist while its intensity is weaker than 34 knots.

\subsection{Profile Quality Analysis}
\label{sec:profile_quality}
Noted that, even with valid wind profiles, there is still a portion of data that has a large difference between the original RMW label and the calculated RMW. 
We consider a wind profile with uncertainty if the distance between origianl RMW label and calculated RMW is more than two standard deviations (fig. 2b). \cref{fig:dataset}(a) demonstrates good and uncertain examples. The green line shows a good TC wind profile, with $V_{max}$at maximum wind speed, RMW at the radius of $V_{max}$, and $R_{34}$ at the radius of wind speed equals to 34 knots. In contrast, the brown line is a profile with uncertainty. Although $V_{max}$ and $R_{34}$ always fit the best-track data, the calculated RMW moves outward a lot, and the calculated outer wind speed may be over-estimated.

As shown in \cref{fig:dataset}(b), the RMW of most of the samples shifts slightly to fit the wind model. We can see that there are 91.6\% of data with RMW difference within two standard deviations (17.4km). Most of the data having significant differences are weak TCs, since that the parametric wind model is developed upon mature TCs.

While the wind profile label we calculated could hardly be perfect, especially with the assumption that the TCs are perfectly symmetric, we still believe that these labels are valuable in tackling the important topic of TC structure.

\section{Proposed Method}
\label{sec:method}

\begin{figure}[t]
\centering
\includegraphics[width=\linewidth]{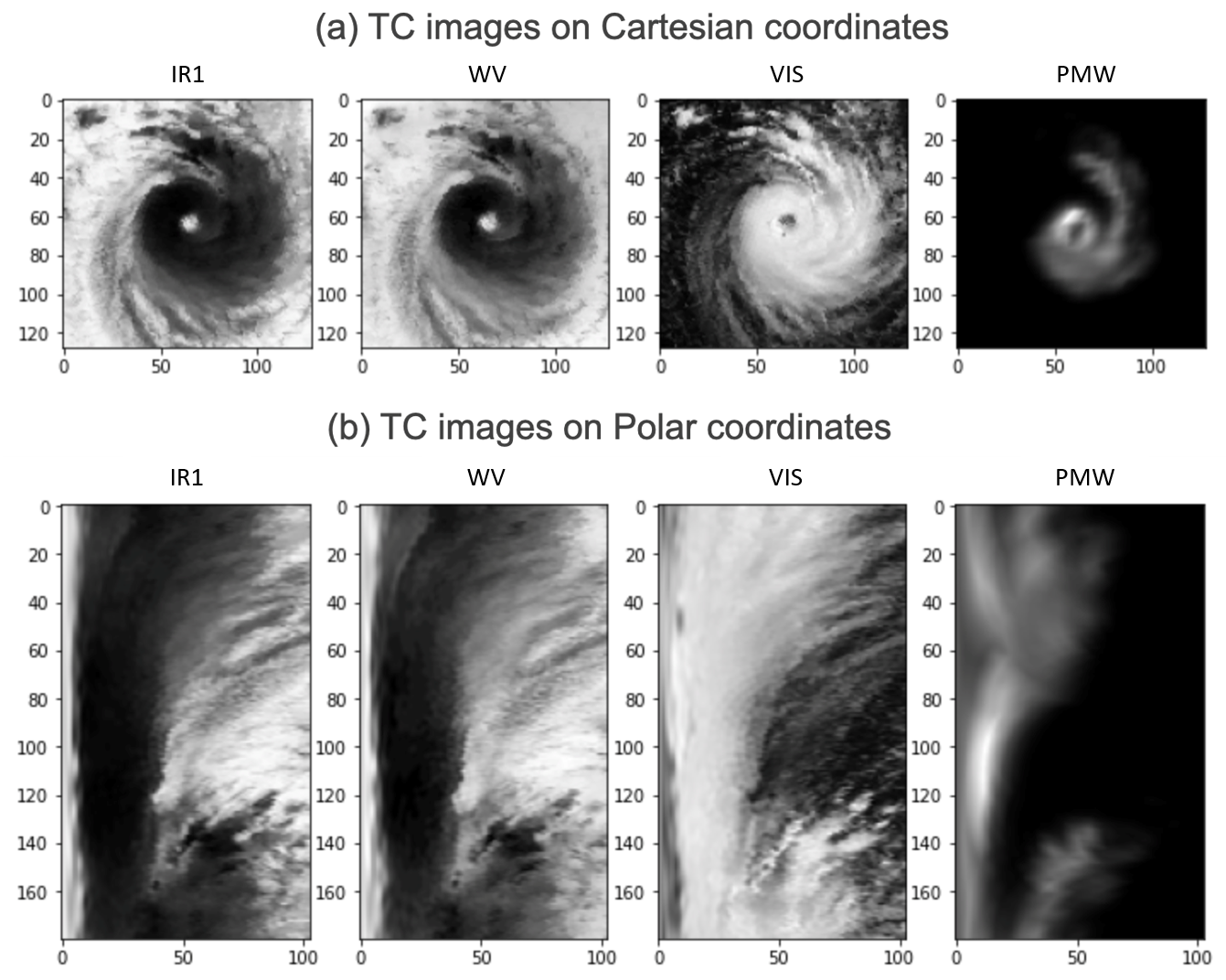}
\caption{Selected TC images on (a) Cartesian coordinates and (b) polar coordinates.}
    \label{fig:coo_example}
\end{figure}

\begin{figure}[t]
\centering
\includegraphics[width=\linewidth]{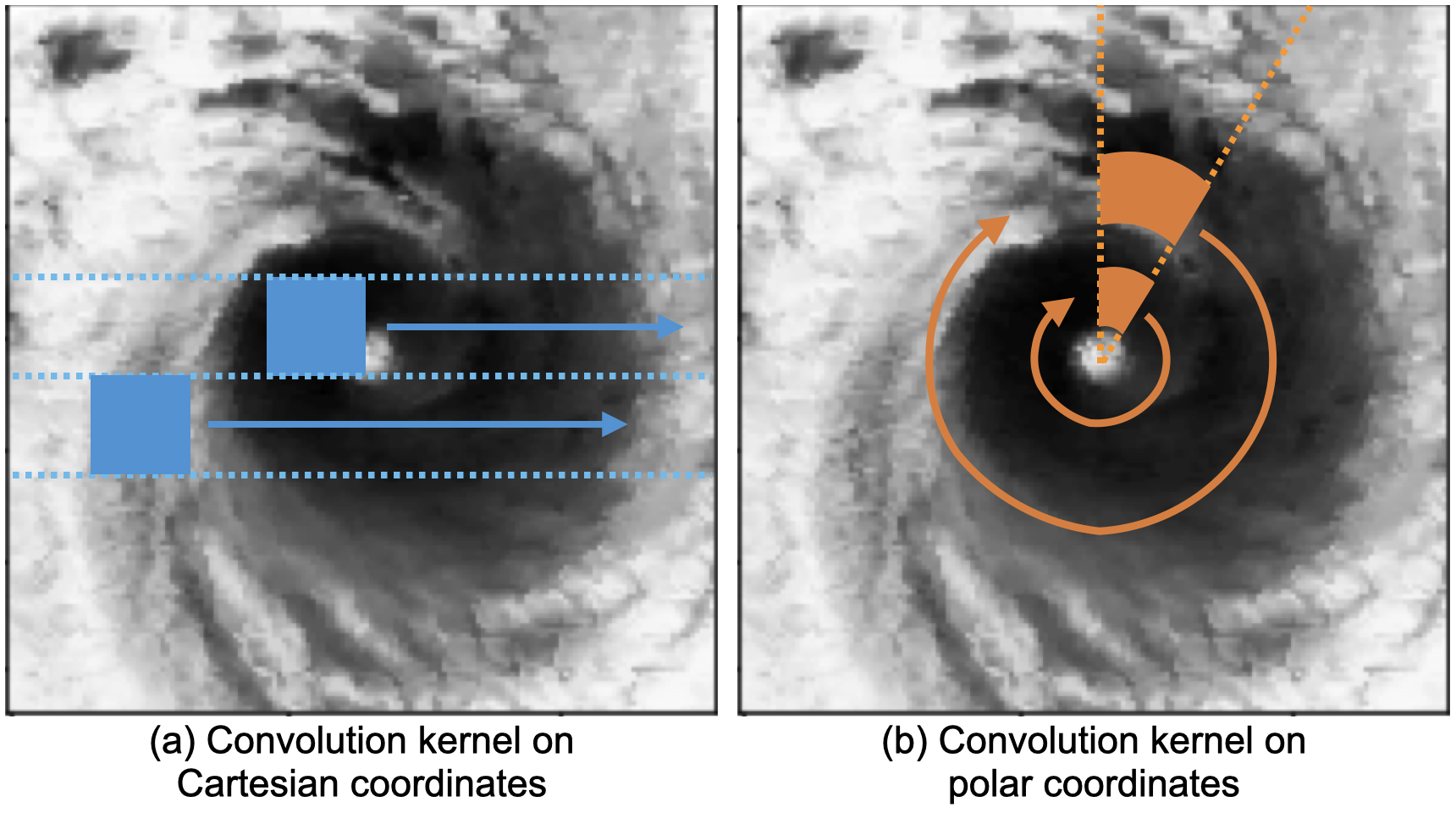}
\caption{A schematic showing convolution kernel working on (a) Cartesian coordinates and (b) polar coordinates.}
	\label{fig:kernel_on_coo}
\end{figure}

According to TCs' spiral nature, a TC  is generally axis-symmetric or point symmetric with the center. Therefore, we propose a unique CNN model that operates on polar coordinates with respect to the TC center.
Before the training, we project all the TC images, originally 128x128 on Cartesian coordinates, to 180x103 images on polar coordinates ( \cref{fig:coo_example}).
Using polar coordinates brings us three benefits:

\begin{enumerate}
    \item They provide more explainable dimensions than those on Cartesian coordinates, allowing us to interpret the model better. Each index in the first dimension (180 points) represents 2 degrees of the directional angle, while each point in the second dimension (103 points) represents 5 kilometers of the radius.
    \item As proposed in \cite{chen2018rotation}, the spiral characteristic of TCs enables us to obtain better results by blending the predictions of an image rotated with several different angles. The effectiveness of this method is also supported in the following work \cite{chen2019estimating}. To rotate an image on Cartesian coordinates, it requires interpolations and probably cropping if we don't want black corners. But on polar coordinates, the only thing we need to do is to roll the image.
    \item On polar coordinates, the meaning of a convolution kernel is a sector, instead of a square, with its vertex pointing to the TC center. The sector mask can further highlight the spiral structure that grows outward from the cyclone eye \cref{fig:kernel_on_coo}. We will discuss the efficiency of convolution masks in different coordinate systems and different shapes later in \cref{sec:kernel_result}.
\end{enumerate}

We stack IR1 and PMW (2 out of all 4 channels), into 180x103x2 images before we pass them into our CNN model. The selection of IR1 and PMW is proven to be the best in \cite{chen2019estimating}.

\subsection{Profiling a TC}
\label{sec:profiling}

\begin{figure}[t]
\centering
\includegraphics[width=\linewidth]{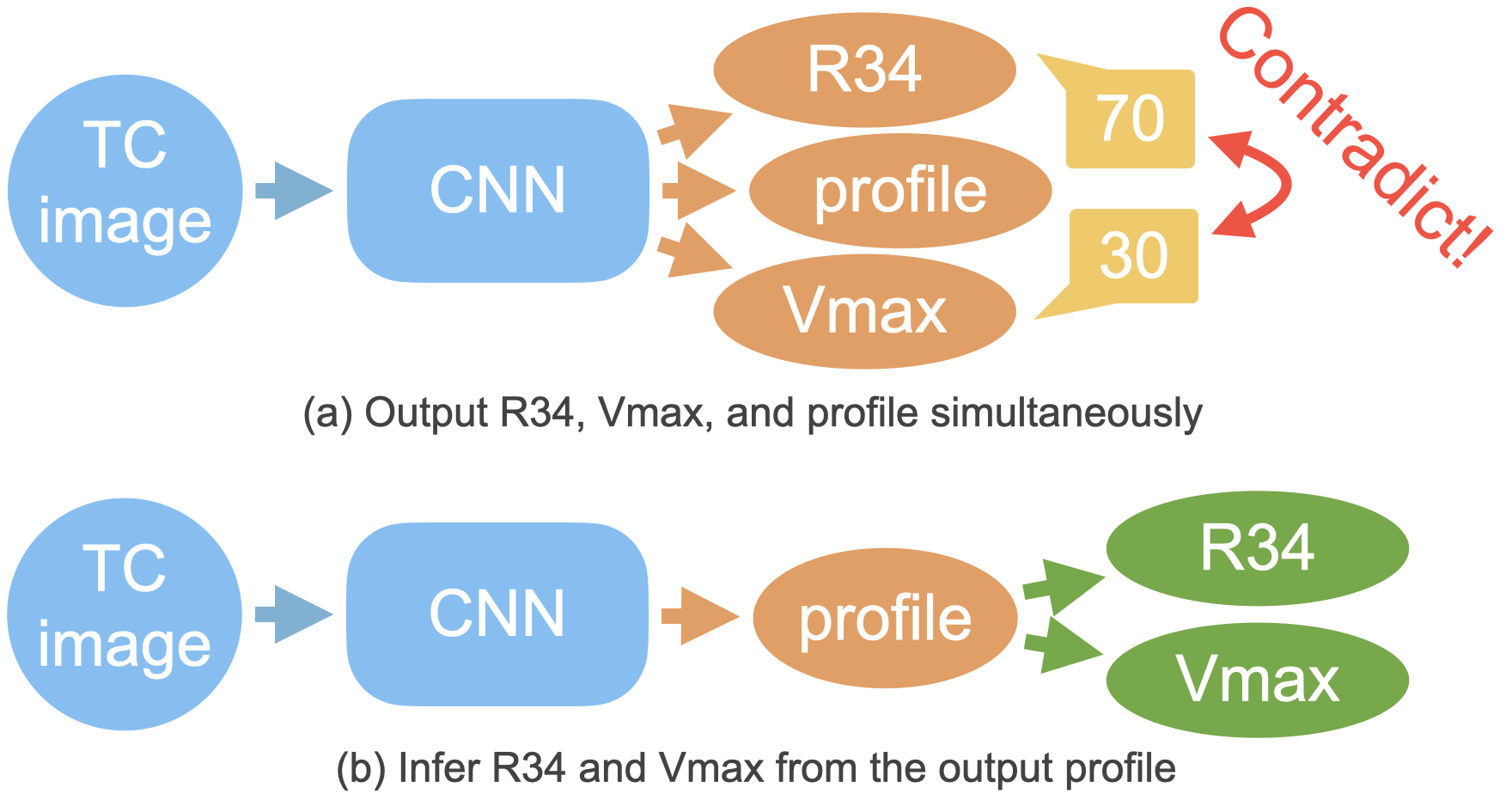}
\caption{A schematic showing two different way to obtain $V_{max}$ and $R_{34}$ from the model. Since method (a) might cause contradictory result, method (b) is recommended.}
	\label{fig:model_flow}
\end{figure}

As suggested in \cref{sec:related}, TC structure is conventionally represented by several parameters: $V_{max}$, RMW, and $R_{34}$. In this work, we hope to further predict the entire wind profile (fig. 2a). Such a wind profile covers the information provided by all the above parameters and provides a more concrete concept of a TC's structure.

However, considering the simplicity and the convenience to compare with other works, we hope that the model can also output $V_{max}$ and $R_{34}$ in addition to the wind profile.
% Besides, their MSE can be added to the loss function for better training.

On the other hand, as mentioned in \cref{sec:data}, only 46\% of the data have profiles. Meanwhile, if the $V_{max}$ of a TC is lower than 34, its $R_{34}$ will naturally be 0. In other words, while every data is guaranteed to have $V_{max}$, not every data has a profile and an $R_{34}$ label.

Therefore, in order to make good use of each data, the loss of $V_{max}$ prediction is also added to the loss function during training. Since the data always has a $V_{max}$ label, we can ensure that there's a loss to be optimized for each data, even if the data don't have profiles and $R_{34}$ labels. To output $V_{max}$ and $R_{34}$ along with the profile, we have two approaches:

A naive way is to let the model output three predictions at the same time: $V_{max}$, $R_{34}$, and the profile. Nevertheless, even if the three outputs share most of the layers, there may be contradictory results. For example, as shown in \cref{fig:model_flow}(a), while $V_{max}$ is lower than 34, the model output a nonzero $R_{34}$.

It is worth noting that there are direct links between the profile and the other two labels. Thus, we suggest to first obtain the profile before inferring $R_{34}$ and $V_{max}$ from the determined profile, as shown in \cref{fig:model_flow}(b).

In the following section, a profile will be denoted as $p$ while the i-th element in the profile will be denoted as $p_i$.

\subsubsection{Inferring $V_{max}$ ($\mathcal{F}_{V}$)}
By definition, $V_{max}$ is the maximum wind speed in the profile, which can be calculated simply by the transformation $\mathcal{F}_{V}$:
\begin{equation}
    \mathcal{F}_{V}(p) = \smash{\displaystyle\max_{i}}(p_i)
\end{equation}

\subsubsection{Inferring $R_{34}$ ($\mathcal{F}_{R}$)}
We first get the biggest index where the wind speed in the profile is greater than 34, and, since the distance between each point is 5 kilometers, multiply the index by 5 to obtain the inferred $R_{34}$ (in km) from the profile.
\begin{equation}
    \mathcal{F}_{R}(p) = \smash{\displaystyle\max_{i}} (i \times [p_i \ge 34]) \times 5
\end{equation}

\subsection{Training Objective}
\label{sec:loss}

For a batch of the data $\mathcal{X}$ and our model $\mathcal{M}$, we first obtain wind profiles using the model,
\begin{equation}
    \mathcal{M}(\mathcal{X}_j) = \overline{\mathcal{P}}_j
\end{equation}

where $\mathcal{X}_j$ and $\overline{\mathcal{P}}_j$ stand for the j-th data and the j-th profile prediction in the batch respectively.

\subsubsection{Profile loss (${l}_{\mathcal{P}}$)}
We calculate point wise mean square error (MSE) between the output profile and the profile label $\mathcal{P}$, the loss will be:
\begin{equation}
{l}_{\mathcal{P}} =
    \smash{\displaystyle\sum_{j}} \smash{\displaystyle\sum_{i}}
    \frac{
        (\overline{\mathcal{P}}_{ji} - \mathcal{P}_{ji})^2
    }{
        i \times j
    }
\label{loss:profile}
\end{equation}

Noted that only when the data have profile label will the profile loss be optimized.

\subsubsection{Intensity loss (${l}_{V_{max}}$)}
We first inferred $V_{max}$ prediction from the profile prediction using transformation $\mathcal{F}_{V}$, then calculate MSE between the $V_{max}$ prediction and the $V_{max}$ label $\mathcal{V}$:
\begin{equation}
{l}_{V_{max}} =
    \smash{\displaystyle\sum_{j}}
    \frac{
        (\mathcal{F}_{V}(\overline{\mathcal{P}}_{j}) - \mathcal{V}_{j})^2
    }{
        j
    }
\label{loss:$V_{max}$}
\end{equation}

\subsubsection{Size loss (${l}_{R_{34}}$)}
We inferred $R_{34}$ prediction from the profile prediction using transformation $\mathcal{F}_{R}$ before calculating MSE between the $R_{34}$ prediction and the $R_{34}$ label $\mathcal{R}$:
\begin{equation}
{l}_{R_{34}} =
    \smash{\displaystyle\sum_{j}}
    \frac{
        (\mathcal{F}_{R}(\overline{\mathcal{P}}_{j}) - \mathcal{R}_{j})^2
    }{
        j
    }
\label{loss:$R_{34}$}
\end{equation}

Finally, as mentioned in \cref{sec:profiling}, we are optimizing ${l}_{\mathcal{P}}$, ${l}_{V_{max}}$, and ${l}_{R_{34}}$ simultaneously. The loss functions are formulated as below:
\begin{equation}
\mathcal{L} = {l}_{\mathcal{P}} + \alpha \times {l}_{V_{max}} + \beta \times {l}_{R_{34}}
\label{loss:overview}
\end{equation}

$\alpha$ and $\beta$ are the factor of intensity loss and size loss, respectively, The factors used for the experiments are provided in \cref{tab:loss_combination}. The detail of the model structure including every layers and blending methods are listed in the appendix. 

\section{Experiments and Analysis}
\label{sec:experiment}

In this section, our attempts in convolution kernel sizes and loss functions are provided first. After that, we look into several actual cases before we compare our proposed model's performance to those of the competitive models.

All models are trained with 2004-2014 TCs, validated with 2015-16 TCs and tested with 2017-2018 TCs.

\subsection{Kernel Size Experiments}
\label{sec:kernel_result}

\begin{table*}[t]
\centering
\begin{tabular}{@{}c|ccccc@{}}
\toprule
Coordinate & (angle, radial) & Profile RMSE (knots) & $V_{max}$ RMSE (knots) & $R_{34}$ RMSE (km) & Selected Epoch \\ \midrule
 & (2, 2) & 14.85 & 12.51 & 70.07 & 20 \\
 & (3, 3) & 14.71 & {\color[HTML]{FE0000} 10.66} & 70.55 & 35 \\
\multirow{-3}{*}{Cartasian} & (4, 4) & 14.89 & 11.32 & 69.27 & 20 \\ \midrule
 & (2, 2) & 14.92 & 11.35 & 68.15 & 60 \\
 & (2, 3) & 14.88 & {\color[HTML]{F56B00} 10.78} & {\color[HTML]{FE0000} 66.33} & 60 \\
 & (2, 4) & 14.67 & 12.24 & 67.09 & 35 \\
 & (3, 2) & 14.53 & 12.08 & 69.33 & 30 \\
 & (3, 3) & 14.63 & {\color[HTML]{F56B00} 11.00} & {\color[HTML]{F56B00} 66.70} & 45 \\
 & (3, 4) & 14.82 & 11.21 & 68.79 & 55 \\
 & (4, 2) & 14.43 & 11.16 & 66.71 & 20 \\
 & (4, 3) & {\color[HTML]{F56B00} 14.28} & {\color[HTML]{F56B00} 11.07} & {\color[HTML]{F56B00} 66.48} & 40 \\
 & (4, 4) & 14.45 & 11.76 & 70.62 & 40 \\
 & (6, 3) & {\color[HTML]{F56B00} 14.21} & 11.22 & 69.49 & 55 \\
\multirow{-11}{*}{Polar} & (8, 3) & {\color[HTML]{FE0000} 13.84} & 12.15 & 70.97 & 55 \\ \bottomrule
\end{tabular}
\caption{The comparison between different kernel shapes. Since the scores vibrate, we select the epoch based on the profile RMSE on the validation data. (4, 3) is selected as the shape of the convolution layers in the final model.}
\label{tab:kernel_shape}
\end{table*}

Since the proposed model is designed to be used in polar coordinates, the shape of the convolution kernel has a more specific meaning. We experimented with the performance of convolution kernels of different shapes.
For simplification, in every model, we use the same strides and the same number of convolution layers. Moreover, each convolution layers in a single model share one kernel shape. For better performance, one can mix different kernel shapes and strides in a model, but considering the simplicity, this is beyond the scope of this work.

\Cref{tab:kernel_shape} shows the performance of different convolution kernels. The experimental result also shows that images on Cartesian coordinates provide decent $V_{max}$ estimates but fall short of predicting profile and $R_{34}$.

Meanwhile, we can observe that the performance on estimating $V_{max}$ and $R_{34}$ is related to the kernel's coverage on the radius. Experimental result shows that choosing a kernel that covers 3 grids on the radius performs best.
In contrast, predicting the profile is more related to how large the angle covered by the kernel is.
As the angle covered is larger, the performance of predicting profile will be better.
However, we also found that as the angle covered becomes larger,
the model is easier to over-fit and thus the accuracy of predicting $V_{max}$ and $R_{34}$ is damaged. In the end, we choose (4, 3) as our kernel shape in the proposed model.

\subsection{Loss Function Combinations}
\label{sec:loss_combination}

\begin{table*}[t]
\centering
\begin{tabular}{@{}lllllll@{}}
\toprule
Loss & $\alpha$ & $\beta$ & Profile RMSE (knots) & $V_{max}$ RMSE (knots) & $R_{34}$ RMSE (km) & Selected Epoch \\ \midrule
Profile & 0 & 0 & 15.93 & 13.55 & 76.95 & 15 \\
Profile+$R_{34}$ & 0 & 0.1 & 15.87 & 13.19 & 72.74 & 35 \\
Profile+$V_{max}$ & 0.3 & 0 & {\color[HTML]{FE0000} 14.18} & 11.32 & 70.60 & 65 \\
Profile+$V_{max}$+$R_{34}$ & 0.3 & 0.1 & 14.37 & {\color[HTML]{FE0000} 11.31} & {\color[HTML]{FE0000} 69.68} & 30 \\ \bottomrule
\end{tabular}
\caption{The comparison between different factor combinations in the loss function. While adding $R_{34}$ loss into loss function provide limited improvement, optimizing $V_{max}$ at the same time help the model learn much better. $\alpha$ and $\beta$ stand for the coefficients mentioned in \cref{loss:overview}. The performance is calculated with the validation data.}
\label{tab:loss_combination}
\end{table*}

We then compare the performance of various combinations of loss functions. As mentioned in \cref{sec:profiling}, we hope that the proposed model can provide high-quality profile, $V_{max}$, and $R_{34}$ predictions at the same time.
\Cref{tab:loss_combination} lists the performance of various combinations of the above three goals in the loss function. The alpha and beta here correspond to the coefficients mentioned in \cref{loss:overview}.

We can observe that, comparing to the model only optimizing $l_{\mathcal{P}}$, the model with additional $l_{V_{max}}$ in the loss function received a decent improvement. In other words, guiding the model to draw the highest point in the curve at the correct height provides a clear direction for the model to do better in fitting the whole line, thus greatly enhanced the performance of the model to predict not only the $V_{max}$ but also the profile and $R_{34}$.

In contrast, adding $R_{34}$ to the loss function hardly improves the model. Our explanation is that for the CNN model, the point where velocity equals 34 in the curve, comparing to the highest point in the curve, is very difficult to grasp. Therefore, it is just better to concentrate on fitting the profile curve and let the $R_{34}$ fit naturally.

According to the above results, we combine $l_{\mathcal{P}}$ and $l_{V_{max}}$ into our loss function of the proposed model.

\subsection{Case Study}
\begin{figure*}[t]
\centering
\includegraphics[width=\linewidth]{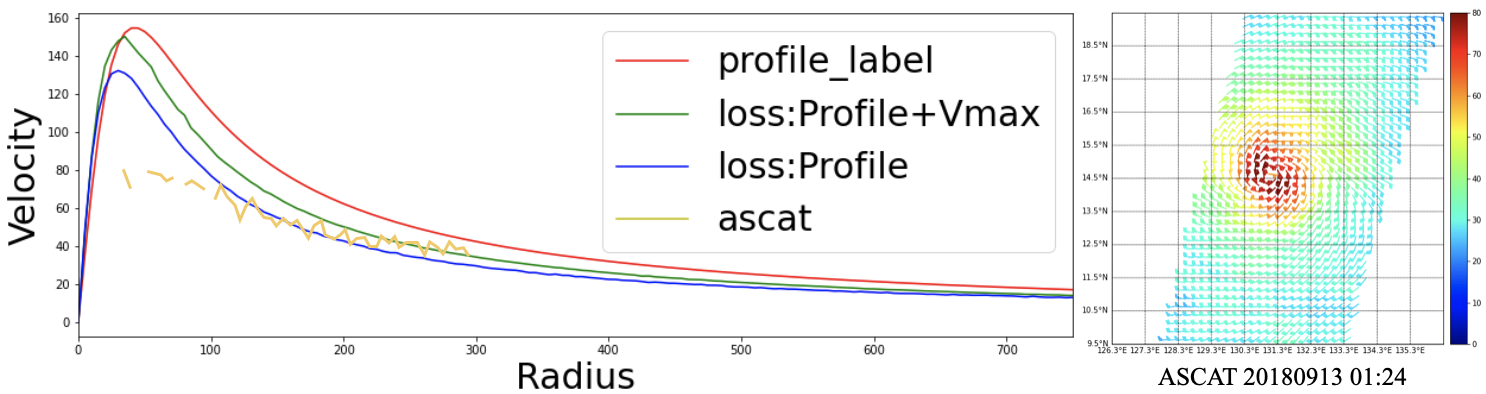}
\caption{Comparing the predicted profiles based on different loss functions (green and blue lines) to the profile label (red line) and the ASCAT observation (gold line, corresponding to the right panel). This figure is generate with the testing data.}
\label{fig:case}
\end{figure*}

\Cref{fig:case} shows a representative case in which we compare the profile label with the prediction of (1) our best model, (2) a model optimized profile loss only, and (3) the ASCAT observation.
From the line chart, we can found that by adding the $l_{V_{max}}$ into the loss function, the model did better fitting the peak ($V_{max}$) of the predicted profile.

On the other hand, the ASCAT observations are restricted by the device limitation. Therefore, when the wind speed is very high (i.e., in the TC inner-core), the ASCAT tend to under-estimate the wind speed. 
In this case (\Cref{fig:case}), our model (green line) did a good job in both accurately estimating the high winds in the inner-core compared to the best-track $V_{max}$ (i.e., the max of the red line) and adequately estimating the TC outer wind comparing the ASCAT profile within 100-300 km radius.

\textit{More cases and interesting observations will be provided in the appendix (and github after the double-blind review).}

\subsection{Performance}
\label{sec:performance}

\begin{table*}[t]
\centering
\includegraphics[width=\linewidth]{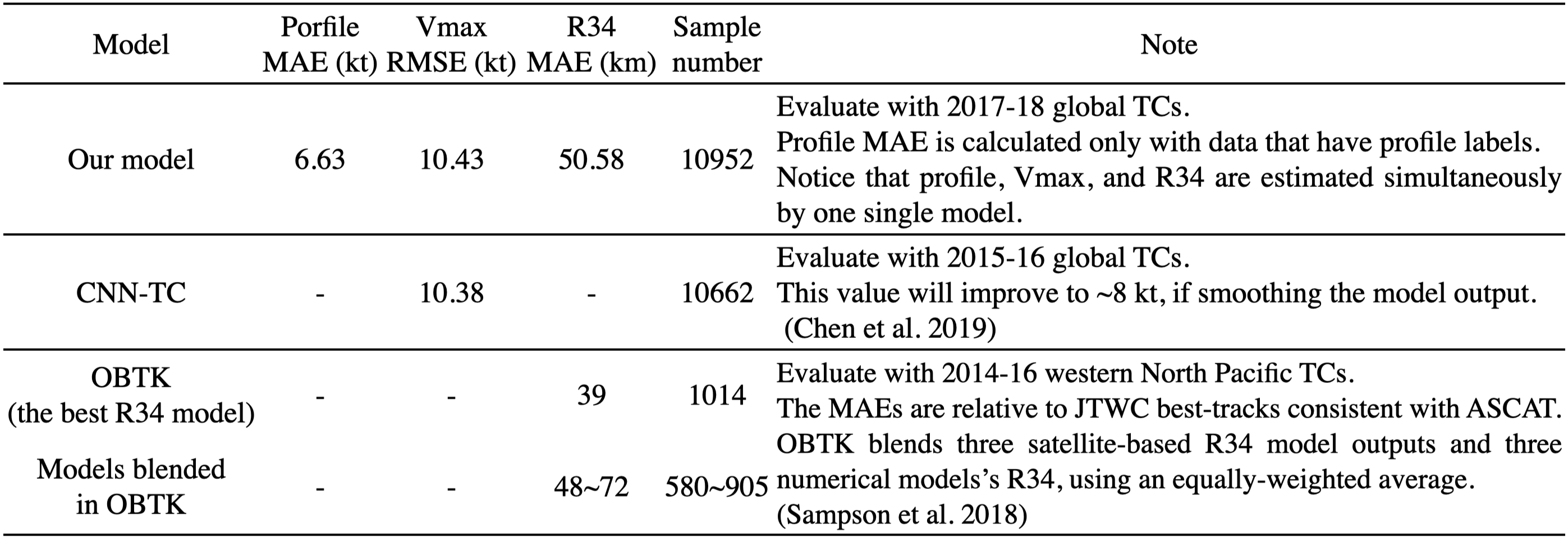}
\caption{Comparing our model to the state-of-the-art methods in both intensity and size estimation.}
\label{tab:performance}
\end{table*}

\Cref{tab:performance} shows the comparison between our proposed model and the state-of-the-art models in TC intensity and size estimation, respectively.

In estimating intensity, the model proposed by \cite{chen2019estimating} is the state-of-the-art in our best knowledge.
By smoothing the output, its performance can be further improved.
For the sake of fairness, we compare the performance without smoothing the output.
If necessary, smoothing techniques can also be applied to our proposed CNN model for better performance.

In estimating TC size, the model of \cite{sampson2018tropical} obtained the best results by blending six independent models.
These six models have their pros and cons and are not always available.
However, simply blending the available estimates of these models with an equally-weighted average leads to better performance. 
On the other hand, our model can not only systematically estimate the TC size, but also be comparable in performance to the best single model \cite{sampson2018tropical} used for blending.

The comparison results suggest that our proposed model can simultaneously predict $V_{max}$ and $R_{34}$ and has comparable performance to state-of-the-art techniques. Moreover, our model provides the radial wind profile, giving us a more concrete concept of TC structure that no other model can provide.

\section{Conclusion}
\label{sec:conclusion}

This paper focuses on an influential but undeveloped task: systematically analyzing the TC structure in terms of its entire radial wind profile.
An organized new dataset with valuable labels for this task is published to facilitate data scientists in the following researches.
By developing on polar coordinates instead of ordinary Cartesian coordinates, we proposed a specialized CNN model that uses rectangular convolution kernels instead of standard square kernels.
We also discovered that optimizing the loss of both intensity estimation and structure estimation at the same time improved our model decently.

With a properly designed model structure and a delicate-composed loss function, our proposed model provides comparable predictions of a TC's size, intensity, and wind profile simultaneously. Most importantly, the prediction is achieved systematically and objectively by using high-availability data, which leads to a more reliable and timely (every 3 h compared to longer than 6 h before) TC forecasting system.

% \begin{ack}
% Use unnumbered first level headings for the acknowledgments. All acknowledgments
% More information about this disclosure can be found at: \url{https://neurips.cc/Conferences/2020/PaperInformation/FundingDisclosure}.

% Do {\bf not} include this section in the anonymized submission, only in the final paper. You can use the \texttt{ack} environment provided in the style file to autmoatically hide this section in the anonymized submission.
% \end{ack}

\bibliography{Chen}

\begin{thebibliography}{23}
\providecommand{\natexlab}[1]{#1}
\providecommand{\url}[1]{\texttt{#1}}
\providecommand{\urlprefix}{URL }
\expandafter\ifx\csname urlstyle\endcsname\relax
  \providecommand{\doi}[1]{doi:\discretionary{}{}{}#1}\else
  \providecommand{\doi}{doi:\discretionary{}{}{}\begingroup
  \urlstyle{rm}\Url}\fi

\bibitem[{Bai, Chen, and Lin(2019)}]{bai2019attention}
Bai, C.-Y.; Chen, B.-F.; and Lin, H.-T. 2019.
\newblock Attention-based Deep Tropical Cyclone Rapid Intensification
  Prediction.
\newblock \emph{arXiv preprint arXiv:1909.11616} .

\bibitem[{Bender et~al.(2016)Bender, Morin, Emanuel, Knaff, Sampson, Ginis, and
  Thomas}]{bender2016impact}
Bender, M.; Morin, M.; Emanuel, K.; Knaff, J.; Sampson, C.; Ginis, I.; and
  Thomas, B. 2016.
\newblock Impact of storm structure and the environmental conditions in the
  rapid intensification of Hurricanes Katrina and Patricia.
\newblock In \emph{32nd Conf. on Hurricanes and Tropical Meteorology}.

\bibitem[{Chan and Chan(2012)}]{chan2012size}
Chan, K.~T.; and Chan, J.~C. 2012.
\newblock Size and strength of tropical cyclones as inferred from QuikSCAT
  data.
\newblock \emph{Monthly weather review} 140(3): 811--824.

\bibitem[{Chen, Chen, and Lin(2018)}]{chen2018rotation}
Chen, B.; Chen, B.-F.; and Lin, H.-T. 2018.
\newblock Rotation-blended CNNs on a new open dataset for tropical cyclone
  image-to-intensity regression.
\newblock In \emph{Proceedings of the 24th ACM SIGKDD International Conference
  on Knowledge Discovery \& Data Mining}, 90--99.

\bibitem[{Chen et~al.(2019)Chen, Chen, Lin, and Elsberry}]{chen2019estimating}
Chen, B.-F.; Chen, B.; Lin, H.-T.; and Elsberry, R.~L. 2019.
\newblock Estimating tropical cyclone intensity by satellite imagery utilizing
  convolutional neural networks.
\newblock \emph{Weather and Forecasting} 34(2): 447--465.

\bibitem[{Demuth et~al.(2004)Demuth, DeMaria, Knaff, and
  Vonder~Haar}]{demuth2004evaluation}
Demuth, J.~L.; DeMaria, M.; Knaff, J.~A.; and Vonder~Haar, T.~H. 2004.
\newblock Evaluation of Advanced Microwave Sounding Unit tropical-cyclone
  intensity and size estimation algorithms.
\newblock \emph{Journal of Applied Meteorology} 43(2): 282--296.

\bibitem[{Figa-Salda{\~n}a et~al.(2002)Figa-Salda{\~n}a, Wilson, Attema,
  Gelsthorpe, Drinkwater, and Stoffelen}]{figa2002advanced}
Figa-Salda{\~n}a, J.; Wilson, J.~J.; Attema, E.; Gelsthorpe, R.; Drinkwater,
  M.~R.; and Stoffelen, A. 2002.
\newblock The advanced scatterometer (ASCAT) on the meteorological operational
  (MetOp) platform: A follow on for European wind scatterometers.
\newblock \emph{Canadian Journal of Remote Sensing} 28(3): 404--412.

\bibitem[{Holland and Merrill(1984)}]{holland1984dynamics}
Holland, G.~J.; and Merrill, R.~T. 1984.
\newblock On the dynamics of tropical cyclone structural changes.
\newblock \emph{Quarterly Journal of the Royal Meteorological Society}
  110(465): 723--745.

\bibitem[{Knaff et~al.(2011)Knaff, DeMaria, Molenar, Sampson, and
  Seybold}]{knaff2011automated}
Knaff, J.~A.; DeMaria, M.; Molenar, D.~A.; Sampson, C.~R.; and Seybold, M.~G.
  2011.
\newblock An automated, objective, multiple-satellite-platform tropical cyclone
  surface wind analysis.
\newblock \emph{Journal of applied meteorology and climatology} 50(10):
  2149--2166.

\bibitem[{Knaff, Longmore, and Molenar(2014)}]{knaff2014objective}
Knaff, J.~A.; Longmore, S.~P.; and Molenar, D.~A. 2014.
\newblock An objective satellite-based tropical cyclone size climatology.
\newblock \emph{Journal of Climate} 27(1): 455--476.

\bibitem[{Knaff et~al.(2016)Knaff, Slocum, Musgrave, Sampson, and
  Strahl}]{knaff2016using}
Knaff, J.~A.; Slocum, C.~J.; Musgrave, K.~D.; Sampson, C.~R.; and Strahl, B.~R.
  2016.
\newblock Using routinely available information to estimate tropical cyclone
  wind structure.
\newblock \emph{Monthly Weather Review} 144(4): 1233--1247.

\bibitem[{Krizhevsky, Sutskever, and Hinton(2012)}]{krizhevsky2012imagenet}
Krizhevsky, A.; Sutskever, I.; and Hinton, G.~E. 2012.
\newblock Imagenet classification with deep convolutional neural networks.
\newblock In \emph{Advances in neural information processing systems},
  1097--1105.

\bibitem[{Matsuoka et~al.(2018)Matsuoka, Nakano, Sugiyama, and
  Uchida}]{matsuoka2018deep}
Matsuoka, D.; Nakano, M.; Sugiyama, D.; and Uchida, S. 2018.
\newblock Deep learning approach for detecting tropical cyclones and their
  precursors in the simulation by a cloud-resolving global nonhydrostatic
  atmospheric model.
\newblock \emph{Progress in Earth and Planetary Science} 5(1): 80.

\bibitem[{Morris and Ruf(2017)}]{morris2017determining}
Morris, M.; and Ruf, C.~S. 2017.
\newblock Determining tropical cyclone surface wind speed structure and
  intensity with the CYGNSS satellite constellation.
\newblock \emph{Journal of Applied Meteorology and Climatology} 56(7):
  1847--1865.

\bibitem[{Powell and Reinhold(2007)}]{powell2007tropical}
Powell, M.~D.; and Reinhold, T.~A. 2007.
\newblock Tropical cyclone destructive potential by integrated kinetic energy.
\newblock \emph{Bulletin of the American Meteorological Society} 88(4):
  513--526.

\bibitem[{Sampson et~al.(2017)Sampson, Fukada, Knaff, Strahl, Brennan, and
  Marchok}]{sampson2017tropical}
Sampson, C.~R.; Fukada, E.~M.; Knaff, J.~A.; Strahl, B.~R.; Brennan, M.~J.; and
  Marchok, T. 2017.
\newblock Tropical cyclone gale wind radii estimates for the western North
  Pacific.
\newblock \emph{Weather and Forecasting} 32(3): 1029--1040.

\bibitem[{Sampson et~al.(2018)Sampson, Goerss, Knaff, Strahl, Fukada, and
  Serra}]{sampson2018tropical}
Sampson, C.~R.; Goerss, J.~S.; Knaff, J.~A.; Strahl, B.~R.; Fukada, E.~M.; and
  Serra, E.~A. 2018.
\newblock Tropical cyclone gale wind radii estimates, forecasts, and error
  forecasts for the western North Pacific.
\newblock \emph{Weather and Forecasting} 33(4): 1081--1092.

\bibitem[{Sampson and Knaff(2015)}]{sampson2015consensus}
Sampson, C.~R.; and Knaff, J.~A. 2015.
\newblock A consensus forecast for tropical cyclone gale wind radii.
\newblock \emph{Weather and Forecasting} 30(5): 1397--1403.

\bibitem[{Tallapragada(2015)}]{tallapragadacoauthors}
Tallapragada, V. 2015.
\newblock Hurricane Weather Research and Forecasting (HWRF) Model: 2015
  Scientific Documentation.
\newblock \emph{NCAR Developmental Testbed Center, Boulder, CO} .

\bibitem[{Velden et~al.(2006)Velden, Harper, Wells, Beven, Zehr, Olander,
  Mayfield, Guard, Lander, Edson et~al.}]{velden2006dvorak}
Velden, C.; Harper, B.; Wells, F.; Beven, J.~L.; Zehr, R.; Olander, T.;
  Mayfield, M.; Guard, C.~C.; Lander, M.; Edson, R.; et~al. 2006.
\newblock The Dvorak tropical cyclone intensity estimation technique: A
  satellite-based method that has endured for over 30 years.
\newblock \emph{Bulletin of the American Meteorological Society} 87(9):
  1195--1210.

\bibitem[{Weatherford and Gray(1988)}]{weatherford1988typhoon}
Weatherford, C.~L.; and Gray, W.~M. 1988.
\newblock Typhoon structure as revealed by aircraft reconnaissance. Part II:
  Structural variability.
\newblock \emph{Monthly Weather Review} 116(5): 1044--1056.

\bibitem[{Wimmers, Velden, and Cossuth(2019)}]{wimmers2019using}
Wimmers, A.; Velden, C.; and Cossuth, J.~H. 2019.
\newblock Using deep learning to estimate tropical cyclone intensity from
  satellite passive microwave imagery.
\newblock \emph{Monthly Weather Review} 147(6): 2261--2282.

\bibitem[{Yang, Lee, and Tippett(2020)}]{yang2020long}
Yang, Q.; Lee, C.-Y.; and Tippett, M.~K. 2020.
\newblock A long short-term memory model for global rapid intensification
  prediction.
\newblock \emph{Weather and Forecasting} 35(4): 1203--1220.

\end{thebibliography}

\iftrue

\clearpage
\appendix

\section{Implementation details}

In the following section, two techniques proposed in the previous work will be explained briefly, including \textbf{auxiliary features} and \textbf{rotation-blending}. Please refer to \citet{chen2019estimating} for more details.

\subsection{Auxiliary Features}
\label{appendix:aux_feature}
In addition to the output from convolution layers, additional features are appended before feeding them into the fully-connected layers.
The auxiliary features are demonstrated to be helpful in improving the precision of estimation \cite{chen2019estimating}. 
These features provide clues such as (1) day of year: stand for seasonal information, (2) local time, and the most influential one: (3) One-hot encoded region codes: region codes is in \{\textit{WPAC, EPAC, CPAC, ATLN, IO, SH}\}, representing 6 different basins.

\subsection{Rotation Blending}
Considering the nature of TCs as a rotating weather system, TC data is rotation invariant.
That is, rotations with respect to the center usually do not affect the estimation of the TC intensity.
\cite{chen2018rotation} demonstrated that the idea of using rotation for augmentation leads to a significant improvement in performance.

During the training phase, each image will be randomly rotated by any degree before feeding into our model.
When it comes to inference, images will be rotated by evenly distributed ten angles ranged from 0 to 360 to collect 10 different estimations. Afterward, these intensity estimations are blended to obtain the final estimate.

Notice that, to rotate images in polar coordinates, we are merely rolling the image upward (\cref{fig:rotation_in_polar}).

\textbf{In this work}, by transforming the images from Cartesian coordinates to polar coordinates, the computing loading is greatly reduced.

\begin{figure}[h]
\centering
\includegraphics[width=\linewidth]{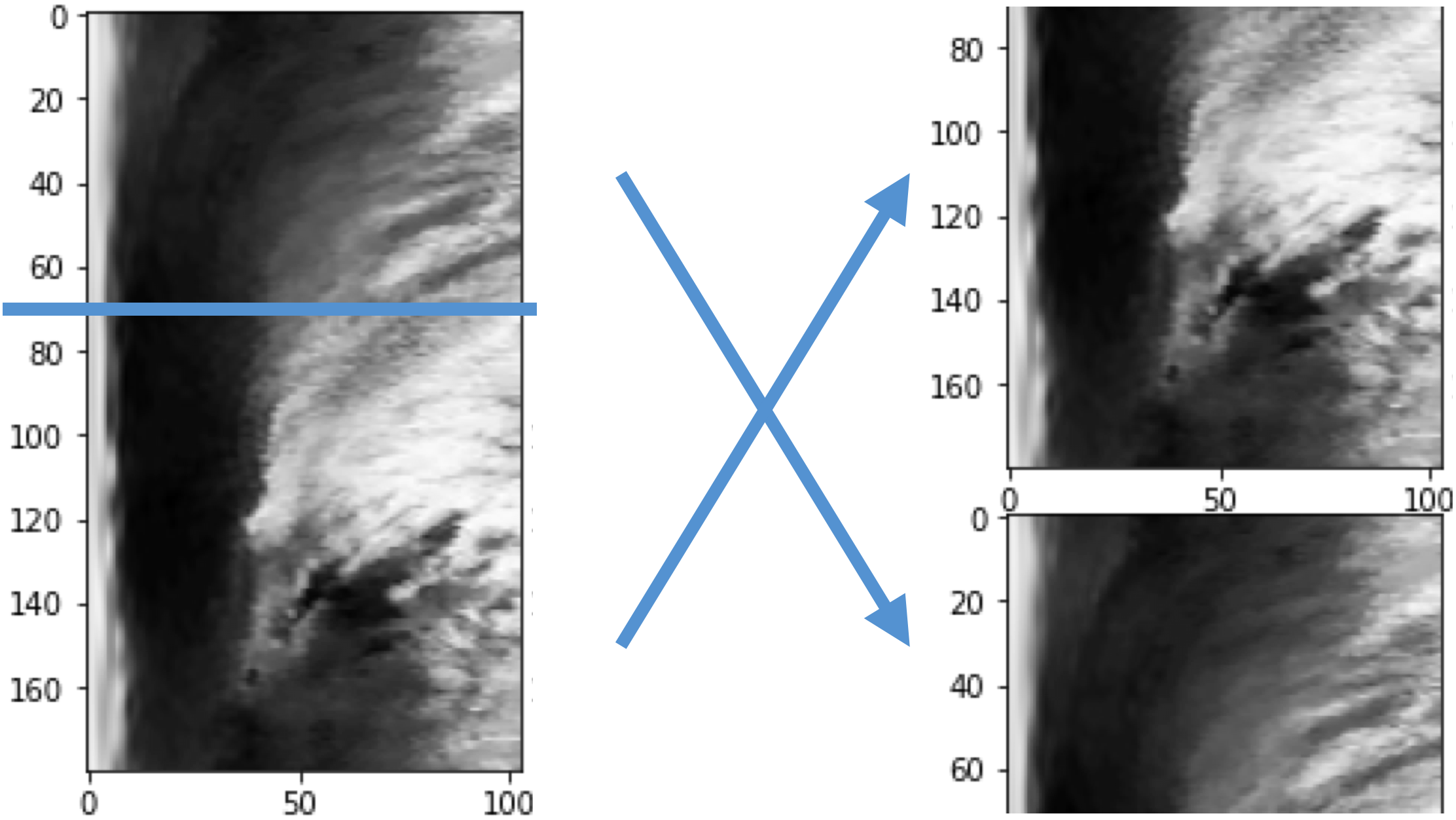}
\caption{A schematic showing how images on polar coordinates are rotated.}
	\label{fig:rotation_in_polar}
\end{figure}

\subsection{Model Structure}

The model structure for the CNN-profiler is detailed in \cref{tab:profiler}.

\begin{table}[ht]
\centering
\begin{tabular}{cccccc}
\hline
\bf Operation & \bf Kernel & \bf Strides & \bf Dim. & \bf BN & \bf activ. \\ \hline
BN & - & - & - & Y & - \\
conv & 4x3 & 2x2 & 16 & Y & relu \\
conv & 4x3 & 2x2 & 32 & Y & relu \\
conv & 4x3 & 2x2 & 64 & Y & relu \\
conv & 4x3 & 2x2 & 128 & Y & relu \\
conv & 4x3 & 2x2 & 256 & Y & relu \\
conv & 4x3 & 2x2 & 512 & Y & relu \\ \hline
\multicolumn{6}{c}{concatenate 10 additional features} \\ \hline
linear & - & - & 256 & Y & relu \\
linear & - & - & 64 & Y & relu \\
linear & - & - & 151 & N & - \\ \hline
\end{tabular}
\caption{Model structure of the CNN-profiler. The first batch normalization layer right serves as z-score normalization. After the convolution layers, 10 dimension features, which were mentioned in \cref{appendix:aux_feature}, are passed into linear layers along with the convolution layers' output.}
\label{tab:profiler}
\end{table}

\section{Extended Case Study}

In \cref{fig:more_case}, we provide 6 cases from different TCs to compare the predicted profiles and the ASCAT observation. Here we have several observations:

\begin{enumerate}
    \item If Vmax loss is added to the loss function, the model tend to 'tap' the maximum velocity with a sharp peak (the middle right case). In contrast, if Vmax loss is not included in the loss function, the model produce much more smooth predictions.
    \item In every cases, the model with additional Vmax loss produce higher curves, which in most cases are more similar to the corresponding profile labels. However, there are still sporadic exceptions (the upper left case).
    \item ASCAT, as the most reliable tropical cyclone size estimation techniques so far, is likely to under-estimate the velocity when the actual velocity is extreme. In contrast, our model provide reliable estimation in both inner and outer core of tropical cyclones.
\end{enumerate}

\begin{figure*}[ht]
\centering
\includegraphics[width=\linewidth]{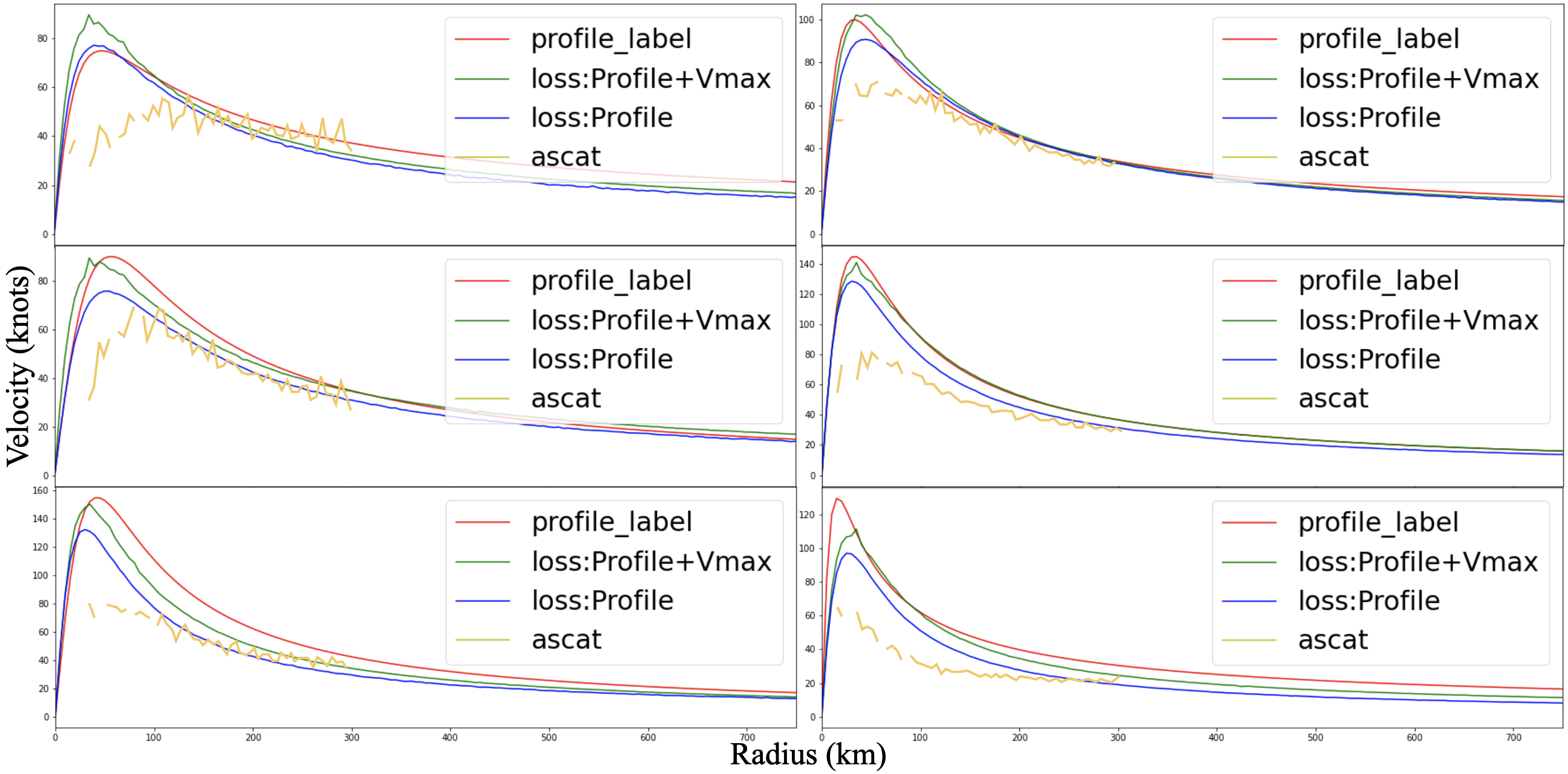}
\caption{An extended version of \cref{fig:case}. This figure compares the predicted profiles based on different loss functions to the profile label and the ASCAT observation using 6 cases from different TCs.}
\label{fig:more_case}
\end{figure*}

\fi

\end{document}